\pdfoutput=1

\documentclass[11pt]{article}

\usepackage{ACL2023}

\usepackage{times}
\usepackage{latexsym}

\usepackage[T1]{fontenc}

\usepackage[utf8]{inputenc}

\usepackage{microtype}

\usepackage{inconsolata}

\usepackage[utf8]{inputenc} 
\usepackage[T1]{fontenc}    
\usepackage{hyperref}       
\usepackage{url}            
\usepackage{booktabs}      
\usepackage{amsfonts}       
\usepackage{nicefrac}       
\usepackage{microtype}      
\usepackage{xcolor}         
\PassOptionsToPackage{hyphens}{url}\usepackage{hyperref}

\usepackage{amssymb}
\usepackage{pifont}

\usepackage{makecell} \usepackage{multirow} \usepackage{mathtools} \usepackage{graphicx} \usepackage{threeparttable} \usepackage{graphicx} \usepackage{svg} \usepackage{amsmath}

\title{Retrieval-Augmented Language Model for Extreme Multi-Label Knowledge Graph Link Prediction}

\author{
  Yu-Hsiang Lin\\
  \texttt{exiled1143@gmail.com} \\
  \And
  {\bf Huang-Ting Shieh} \\
  \texttt{huangtingshieh@gmail.com}\\
  \And
  {\bf Chih-Yu Liu} \\
  \texttt{cyliu0513@gmail.com}\\
  \AND
  {\bf Kuang-Ting Lee} \\
  \texttt{summer51202@gmail.com}
  \And
  {\bf Hsiao-Cheng Chang} \\
  \texttt{student0804@gmail.com} \\
  \AND
  {\bf Jing-Lun Yang} \\
  \texttt{simonhandsome99876@gmail.com}
  \And
  {\bf Yu-Sheng Lin} \\
  \texttt{biolin@cmlab.csie.ntu.edu.tw} \\}

\begin{document}
\maketitle
\begin{abstract}
Extrapolation in Large language models (LLMs) for open-ended inquiry encounters two pivotal issues: (1) hallucination and (2) expensive training costs. These issues present challenges for LLMs in specialized domains and personalized data, requiring truthful responses and low fine-tuning costs. Existing works attempt to tackle the problem by augmenting the input of a smaller language model with information from a knowledge graph (KG). However, they have two limitations: (1) failing to extract relevant information from a large one-hop neighborhood in KG and (2) applying the same augmentation strategy for KGs with different characteristics that may result in low performance. Moreover, open-ended inquiry typically yields multiple responses, further complicating extrapolation. We propose a new task, the extreme multi-label KG link prediction task, to enable a model to perform extrapolation with multiple responses using structured real-world knowledge. Our retriever identifies relevant one-hop neighbors by considering entity, relation, and textual data together. Our experiments demonstrate that (1) KGs with different characteristics require different augmenting strategies, and (2) augmenting the language model's input with textual data improves task performance significantly. By incorporating the retrieval-augmented framework with KG, our framework, with a small parameter size, is able to extrapolate based on a given KG. The code can be obtained on GitHub: \url{https://github.com/exiled1143/Retrieval-Augmented-Language-Model-for-Multi-Label-Knowledge-Graph-Link-Prediction.git}

\end{abstract}

\begin{figure*}
  \centering
  \includegraphics[width=1.5\columnwidth]{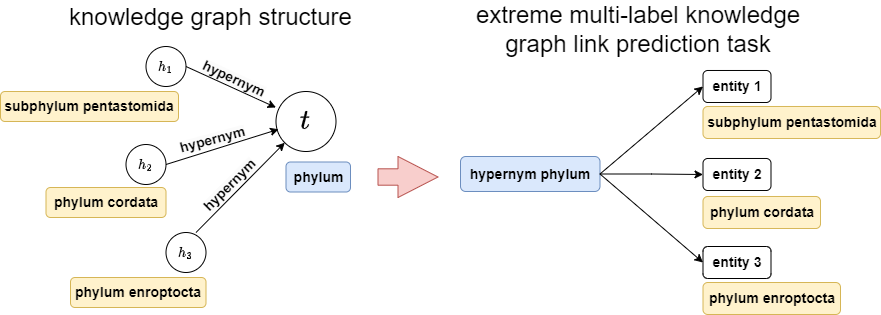}
  \caption{An Illustration of the Extreme Multi-label Knowledge Graph Link Prediction Task. Each $h_i$, $i \in {1, 2, 3}$, denotes the head entity of a triple, and $t$ denotes a tail entity. Consider three triples with the same relation and $t$. This can be reformulated into a multi-label problem by giving relation and tail as input raw text while the $h_{1}$, $h_{2}$, and $h_{3}$ are the corresponding labels.}  
  \label{fig:task_explain}
\end{figure*}

\section{Introduction and Related Works}

We say a language model has hallucinating issues \cite{zhang2023siren, dziri2022origin} if it generates a context that contradicts real-world knowledge. The hallucinating issue and the expensive training or fine-tuning costs pose significant challenges in applying LLMs to specialized domains or personalized data that require high precision and truthful responses. In order to alleviate the hallucination, existing works attempt to augment the input of the language model with information from a knowledge graph \cite{dziri2021neural, ji2023rho, baek2023knowledge}. Moreover, by employing the retrieval augmented framework, the language model is able to have better performance with considerably fewer parameters \cite{izacard2022few, huang2023raven}.

Although the two aforementioned problems are alleviated by incorporating the retrieval-augmented framework and a knowledge graph, previous works still pose two drawbacks. First of all, since the one-hop neighborhood of a node can be large, they may fail to leverage the most relevant one-hop neighbors from the one-hop neighborhood. Moreover, applying the same augmentation strategy to knowledge graphs with different characteristics will lead to low performance.

In real-world extrapolation, a given inquiry typically yields multiple responses. This is similar to a knowledge graph link prediction task. A traditional knowledge graph link prediction task is to (1) predict a tail entity by giving a head entity and a relation or (2) predict a head entity by giving a tail entity and a relation. However, the traditional knowledge graph link prediction task generates only one response for a given inquiry, which may not fit the real-world application. To better conform to real-world situations, we propose a new task called extreme multi-label knowledge graph link prediction. Our proposed task is to predict multiple tail entities by giving a head entity and a relation or to predict multiple head entities by giving a tail entity and a relation.

By incorporating the retrieval-augmented framework with a given knowledge graph and the corresponding textual data, our proposed framework is able to extrapolate based on a given knowledge graph with fewer parameters. Since FB15k-237 and WN18RR are two large knowledge graphs with corresponding textual data, we choose to evaluate our framework on these two datasets. Our experiments demonstrate that (1) knowledge graphs with different characteristics require different augmenting strategies, and (2) the textual data applied to augment the input of the language model significantly improves the performance of our proposed task.

Since the input of a transformer is limited in length, it's better for the textual data on the input side to be as concise as possible. Therefore, we propose a textual dataset that corresponds to FB15k-237. In contrast to the original dataset, the textual data in our proposed dataset is more concise. In addition, a large number of entities in a knowledge graph leads to a high dimensional classification layer of the language model. The model with a high dimensional classification layer may yield poor precision at $k$ when optimized by a binary cross entropy loss. This work proposes a loss function and a training strategy to improve the precision at $k$.

\section{Method}
\label{method1}

\subsection{Problem formulation}
\label{problem_formulation}
A knowledge graph $G=(V, E)$ is a directed entity graph where the set of nodes ($V$) is a set of entities, and the set of edges ($E$) is a set of relations. We can rewrite the knowledge graph as $G \coloneqq \{\,(h, r, t) \mid h, t \in V, \, r \in E  \,\} $
where $(h,r,t)$ is called a triple in which the direction of $r$ (relation) is from $h$ (head) to $t$ (tail).

A knowledge graph link prediction task is to infer the missing entity in a triple $(h, r, t)$. That is, to predict $h$ given $(r, t)$ or $t$ given $(h, r)$. However, there may exist multiple one-hop neighbors for a given node and a given relation. This implies that a given $(h, r)$ may correspond to multiple different tail nodes and a given $(r, t)$ may correspond to multiple different head nodes. Therefore, we formulate this problem into an extreme multi-label knowledge graph link prediction problem. For example, Figure ~\ref{fig:task_explain} shows an illustration of the extreme multi-label knowledge graph link prediction task. Each $h_i$, $i \in {1, 2, 3}$,  denotes a head entity of a triple, and $t=phylun$ denotes a tail entity.The three triples, $(h_1,~hypernym,~t)$, $(h_2,~hypernym,~t)$, $(h_3,~hypernym,~t)$, can be transformed into an input raw text, $\it hypernym~phylun$, with multiple labels, $h_1$, $h_2$, and $h_3$. A head or a tail is an entity. Let $Z$ be the set of all entities including all heads and tails. We collect all such $(h, r)$ and $(r, t)$ in the training triples into a set $\mathcal{E}_{train}$. The set of input raw text for training is defined as $X_{train} \coloneqq \{\,x_{i} \mid x_{i} = (h, r) \in \mathcal{E}_{train} \, \, or \, \, x_{i} = (r, t) \in \mathcal{E}_{train} \,\}$. For each $ x_{i} \in X_{train} $ there exists a $y_{i} \in B^{k}$, where $B=\{ 0,1 \}$ and $k$ is the number of training entities, such that $y_{i}$ is the label of $x_{i}$. Our goal is to learn a function $f(x_{i}) \in B^{k}$ for each $x_{i} \in X_{train}$ which output an one-hot vector $w=(w_{1}, ..., w_{k})=f(x_{i})$ where $w_j =1$ if $z_j \in Z$ corresponds to $x_i$, and $w_j =0$ otherwise.

\begin{figure*}

  \centering
  \includegraphics[width=1.5\columnwidth]{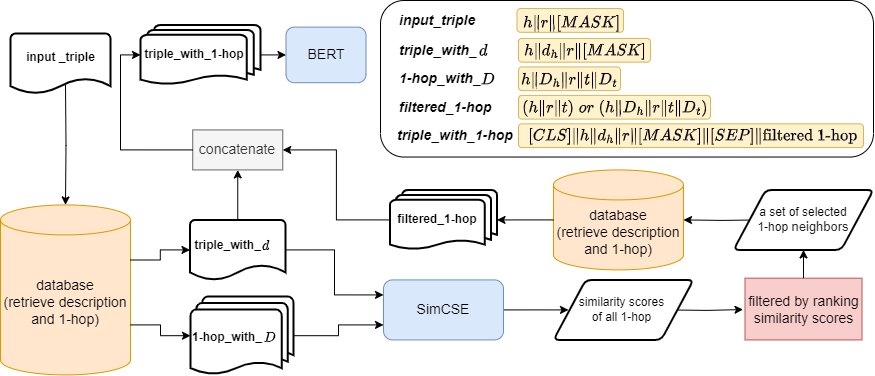}
  \caption{An Illustration of the Proposed Framework. $h$: head entity, $r$: relation, $t$: tail entity, $[CLS]$: [CLS] token, $[MASK]$: [MASK] token, $d$: description of a given node in an incomplete triple, $D$: description of an one-hop neighbor, $d_{h}$: description of the given head entity in an incomplete triple, $D_{h}$: description of the given head entity in an one-hop neighbor, $D_{t}$: description of the given tail entity in an one-hop neighbor, and $\|$: concatenate. This figure shows the format of data at each stage of our framework. Suppose given the \emph{input triple}$\coloneqq h\|r\|[MASK]$, our framework will give out \emph{triple with 1-hop}$\coloneqq [CLS]\|h\|d_{h}\|r\|[MASK]\|[SEP]$$\|$\emph{filtered 1-hop}  as the input raw text of the BERT model.}
  \label{fig:framework}
\end{figure*}

\begin{table}[]

\centering
\scalebox{0.82}{
\begin{tabular}{lrrrr}

\hline
          & $N_{n}$ & $N_{r}$ & $T_{train}$ & $T_{test}$ \\ \hline
FB15k-237 & 40,943   & 237     & 289,605      & 20,439 \\ \hline  
WN18RR    & 14,543   & 11      & 89,869       & 2,947 \\ \hline
 & $\hat{N}_{1\texttt{-}hop}$ & $T_{D}$ &$N_{train}$ & $N_{test}$ \\ \hline 
 FB15k-237 & 31.514 & 78,288 & 155,382      & 22,813  \\ \hline
 WN18RR     &  3.620 & 48,029   & 104,412      & 5,393 \\ \hline
 & $ L$  & $\bar{L}$ & $\hat{L}$ & $S$ \\ \hline
 FB15k-237 &   14,560 & 3.728     & 31.514 &  174,360,664 \\ \hline
WN18RR     &  32,558 & 1.632     & 3.620 & 196,630,257 \\ \hline
  \multicolumn{5}{l}{\makecell{\footnotesize{$N_{n}$: \# nodes, $N_{r}$: \# relations, $T_{train}$: \# training triples, $T_{test}$: }}} \\ 
 \multicolumn{5}{l}{\makecell{\footnotesize{  \# testing triples, $\hat{N}_{1\texttt{-}hop}$: \# 1-hop neighbors per node (train), }}} \\
\multicolumn{5}{l}{\makecell{ \footnotesize{ $N_{train}$: \# training samples, $N_{test}$: \# testing samples, $ L$: \# labels,}}} \\

\multicolumn{5}{c}{\makecell{ \footnotesize{  $\bar{L}$: average labels per sample, $\hat{L}$: average samples per label,  $T_{D}$:}}}\\
\multicolumn{5}{c}{\makecell{\footnotesize{ \# disconnected triples, $S$: \# parameters for the language model. }}}\\
\multicolumn{5}{c}{\makecell{\footnotesize{ Please refer to the SimCSE paper for information on the number }}}\\
\multicolumn{5}{c}{\makecell{\footnotesize{ of parameters used in our retriever model.}}}\\
\end{tabular}}
\caption{The statistics of datasets used in our experiments.}
\label{datasets_statistics}

\end{table}

\subsection{Datasets}
\label{datasets}

We use two large knowledge graphs in our work, namely, WN18RR and FB15k-237. Statistics of each knowledge graph are shown in Table~\ref{datasets_statistics}. Two of the statistics that may increase the difficulty of the proposed task are discussed in the following. One of them is $\hat{N}_{1\texttt{-}hop}$, the number of one-hop neighbors per node. It will be fairly difficult for the model to infer from a node to a target node if this number is small. The other is $T_{D}$, the number of disconnected triples. A triple is considered disconnected if the two nodes in a given triple have no other edges connected to them. This kind of triple is isolated from the graph, making it difficult for a model to infer from a node to such a triple. Note that although the $T_{D}$ of WN18RR is smaller than that of FB15k-237, the proportion of the disconnected triples in WN18RR is much larger than that of FB15k-237 since WN18RR has a smaller number of training triples $T_{train}$. As the objective of the link prediction task is to infer from a given node to another, these two characteristics of WN18RR may increase its difficulty.

 \begin{table*}

\centering
\scalebox{0.72}{
\begin{tabular}{llll}
\hline
 & entity & short description & original description \\ \hline
  FB15k-237 & Bolt & 2008 American computer & Bolt tells the story of a dog who is convinced \\
 & & animated film &   that his role as a super dog is reality. When he \\
  & & &    is ripped from his world of fantasy, and action\\
  & & &   by his own doing. His own obsession  with his\\ 
  & & &   owner and keeping her protected from the green\\
  & & &   eyed man of the television show he works on\\
  & & &    completely absorbs his life. This takes him to\\
  & & &  the point of no return when he believes that she\\
    & & &  has been kidnapped, and he accidentally gets  \\
    & & &  packaged and shipped to New York city in pursuit\\
    & & &  of his owner. This is when the story unfolds, and\\
    & & &  he goes through a transitional period where he \\
    & & & learns that he is as super as every other dog. His\\
    & & & disbelief of his abilities being non existent fuels\\
    & & &  a lot of different emotional changes, and \\
    & & & eventually comes to a reality of who he really is.\\
    & & &  This all culminates when he finds his owner and\\
    & & &  reunites with her in the action packed climax of\\
    & & &  the movie.\\ \midrule
WN18RR & sandarac & durable fragrant & durable fragrant wood; used in building (as in \\
 & &  wood &   the roof of the cathedral at Cordova, Spain) \\ \hline
 \multicolumn{4}{c}{\makecell{ }}
 
 \end{tabular}}
 \caption{The comparison between the original description and the shorter version.}
 \label{des_compare}
 \end{table*}

 \begin{table*}
     \centering

     \scalebox{0.78}{
     \begin{tabular}{crrrr}
     \hline
     \multirow{2}{*}{Datasets of entity descriptions} & \multicolumn{2}{c}{short version} & \multicolumn{2}{c}{original version} \\ \cline{2-5}
      & FB15k-237 & WN18RR & FB15k-237 & WN18RR \\ \hline
      $\bar{N}_{tokens}$ & 6.570 & 10.581 & 189.137 & 17.001 \\ \hline
      \multicolumn{5}{c}{\makecell{ $\bar{N}_{tokens}$ : Average number of tokens}} \\
     \end{tabular}}
     \caption{The comparison of the average lengths of entity descriptions of the two datasets used in our experiments.}
     \label{des_length}
 \end{table*}

A dataset of entity descriptions comprises of the description corresponding to each entity in a knowledge graph. For convenience, we will refer to the dataset of entity descriptions of a knowledge graph as the \emph{description dataset} in the rest of this article. Table~\ref{des_compare} shows examples of entity descriptions of the two knowledge graphs. The description datasets of WN18RR and FB15k-237 are used in the previous works ~\cite{li2022multi, wang2021structure} to help their models to learn better entity embeddings. These works concatenate the triples with the corresponding descriptions and one-hop neighbors as our raw text input. With limited input length of the model, we truncate the description into a shorter version without changing the original meaning to maximize the number of triples in a fixed-length raw text input. For the description dataset of WN18RR, we extract the content before the first semicolon from the original description. As for FB15k-237, we propose a new description dataset based on Wikidata, Wikipedia, and other resources from the internet because the original FB15k-237 description dataset has extremely long descriptions. Table ~\ref{des_compare} shows examples of the short descriptions corresponding to the original descriptions. Table ~\ref{des_length} shows the average length of the descriptions in the two datasets, where the length is measured in terms of the number of tokens using the BERT-base-cased  ~\cite{devlin2018bert} tokenizer.

\subsection{Framework}
\label{sec_framework}

The proposed framework is shown in Figure ~\ref{fig:framework}. The BERT~\cite{devlin2018bert} model is employed as our language model, and the SimCSE~\cite{gao2021simcse} model plays a critical role in the retrieval process.

\textbf{Notation.} Our task is to infer all target tails given a $h$ and a $r$ or to infer all target heads given a $r$ and a $t$. We defined an incomplete triple as $r \| t$ or $h \| r$ where the symbol $\|$ represents string concatenation. Let $d$ be a generic notation that denotes the description of a given node in an incomplete triple. Let $d_{h}$ and $d_{t}$ respectively denote the description of $h$ and $t$ in an incomplete triple. The key idea of the proposed framework is to incorporate the one-hop neighbors of a given node in the incomplete triple. Suppose there are $n$ one-hop neighbors. We defined the triple formed by the given node in the incomplete triple, a neighbor node of the given node, and a relation between the two nodes as an \emph{one-hop~triple}. Let $D$ be a generic notation that denotes the description of a neighbor node. Let $D_{h}$ denote the description of the head entity of an one-hop triple with a given node $h$. Similarly, let $D_{t}$ denote the description of the tail entity of an one-hop neighbor with a given node $t$.

Here, we provide a brief overview of the retrieval process, and the remaining paragraphs in this section will delve into the details. Firstly, \textbf{input\textunderscore triple} is fed into the proposed framework and processed into \textbf{triple\textunderscore with\textunderscore} $\boldsymbol{d}$. Then we concatenate the \textbf{triple\textunderscore with\textunderscore}$\boldsymbol{d}$ with \textbf{filtered\textunderscore 1-hop} into \textbf{triple\textunderscore with\textunderscore 1-hop} in the framework. Finally, the \textbf{triple\textunderscore with\textunderscore 1-hop} is sent into the BERT model as the raw text input. Note that by taking the \textbf{triple\textunderscore with\textunderscore}$\boldsymbol{d}$ and \textbf{1-hop\textunderscore with\textunderscore}$\boldsymbol{D}$ as inputs to the SimCSE model and going through a series of processes, we can obtain \textbf{filtered\textunderscore 1-hop}. 

\textbf{Stage 1 of retrieval process.} The input of the framework is defined as follows:
\begin{equation}
\label{eq:fw1}
\begin{split}
\textbf{input\textunderscore triple} &= [MASK] \| r \| t  \; \text{ or } \; \\
\textbf{input\textunderscore triple} &= h \| r  \|  [MASK]
\end{split}
\end{equation}
where the symbol $\|$ represents string concatenation. The $[MASK]$ token used here has the same function as the $[MASK]$ token in the original BERT model~\cite{devlin2018bert}. The \textbf{input\textunderscore triple} in equation~\ref{eq:fw1} corresponds to the \textbf{input\textunderscore triple} in Figure ~\ref{fig:framework}.

Then by introducing entity descriptions into the \textbf{input\textunderscore triple}, we obtain \textbf{triple\textunderscore with\textunderscore}$\boldsymbol{d}$. The \textbf{triple\textunderscore with\textunderscore}$\boldsymbol{d}$ is defined as follows:
%Followed by LP-BERT~\cite{li2022multi}, entity descriptions corresponding to the entities are introduced to enhance the result. In this work, we concatenate the entity with its corresponding description, and the input raw text in equation~\ref{eq:fw1} can be reformulated as:
\begin{equation}
\label{eq:fw2}
\begin{split}
\textbf{triple\textunderscore with\textunderscore } \boldsymbol{d} &= [MASK] \| r \| t \| d_{t}  \; \text{ or } \;\\  
\textbf{triple\textunderscore with\textunderscore } \boldsymbol{d} &= h \| d_{h} \| r  \|  [MASK]
\end{split}
\end{equation}
where $d_{h}$ and $d_{t}$ are descriptions of $h$ and $t$ respectively. The input format in equation~\ref{eq:fw2} corresponds to the format of \textbf{triple\textunderscore with\textunderscore}$\boldsymbol{d}$ in Figure ~\ref{fig:framework}. To prevent leaking the correct answer, only the description of the tail entity is added while the head entity is masked and vice versa.

\textbf{Stage 2 of retrieval process.} The \textbf{1-hop\textunderscore with\textunderscore}$\boldsymbol{D}$ is obtained by incorporating one-hop triple with its entity descriptions. Due to the length limitation of an input sequence of the BERT, we need to select the most relevant information from these one-hop triples. This is where SimCSE~\cite{gao2021simcse} comes into play. We calculate the similarity between the \textbf{triple\textunderscore with\textunderscore}$\boldsymbol{d}$ and each one-hop triple from the one-hop neighborhood, as shown in Figure ~\ref{fig:framework}. Suppose given $\textbf{triple\textunderscore with\textunderscore } \boldsymbol{d} = [MASK] \| r \| t \| d_{t}$ and a neighbor of $t$, say $q$, with a relation regardless of direction $r_q$ in between. Then the description of $q$ is denoted as $D_{q}$. To ensure that the SimCSE obtains the complete information of each one-hop triple, the  \textbf{1-hop\textunderscore with\textunderscore }$\boldsymbol{D}$ takes the form:
\begin{equation}
\label{eq:fw3}
\begin{split}
\textbf{1-hop\textunderscore with\textunderscore } \boldsymbol{D} & \coloneqq q \| D_{q} \| r_{q} \| t \| d_{t}  \; \text{ or } \;\\ 
\textbf{1-hop\textunderscore with\textunderscore } \boldsymbol{D} & \coloneqq t \| d_{t} \| r_{q} \| q \| D_{q},
\end{split}
\end{equation}
and vice-versa when $\textbf{triple\textunderscore with\textunderscore } \boldsymbol{d} = h \| d_{h} \| r \| [MASK]$. Note that the given $t$ in $\textbf{triple\textunderscore with\textunderscore } \boldsymbol{d} = [MASK] \| r \| t \| d_{t}$ may act as head or tail in an one-hop triple. 
%Inspired by GAT~\cite{velickovic2017graph}, we introduce the one-hop neighborhood to improve the performance of our work. 

\textbf{Stage 3 of retrieval process.} The one-hop neighbors are sorted in descending order according to the similarity scores between \textbf{triple\textunderscore with\textunderscore $d$} and each \textbf{1-hop\textunderscore with\textunderscore $D$}. Then we select a desired number of one-hop triples from the one-hop neighborhood according to the sorted similarity scores. To prevent leaking the correct answer, only the one-hop triples of the tail entity are introduced while the head entity is masked and vice versa.

Due to the different characteristics of these two knowledge graphs, we propose two methods, one for each knowledge graph, for choosing the desired number of one-hop triples. For datasets with low $\hat{N}_{1\texttt{-}hop}$ and a large proportion of disconnected triples such as WN18RR, the chosen one-hop triples must provide as much information as possible. Therefore, we select as many one-hop triples as possible with $[SEP]$ token in between such that the total number of input tokens of the \textbf{triple\textunderscore with\textunderscore 1-hop} does not exceed the maximum input length of the BERT model, which is $512$ tokens in our experiment. The \textbf{filtered\textunderscore 1-hop} of WN18RR takes the form:
\begin{equation}
\label{eq:fw4}
\begin{split}
& h \| r_{1} \| t_{1} \| [SEP]\| ... \| [SEP] \| h \| r_{n} \| t_{n} \; \text{ or } \;\\ 
& h_{1} \| r_{1} \| t \| [SEP]\| ... \| [SEP] \| h_{n} \| r_{n} \| t
\end{split}
\end{equation}
The former is for the tail-masked version, and the latter is for the head-masked version.

For the FB15k-237 dataset, we found that most entities are proper nouns, such as names of films or celebrities, and this is where the descriptions come into play. We select the top three one-hop triples according to similarity scores and concatenate each one-hop triple with its descriptions of the two nodes in the one-hop triple. The \textbf{filtered\textunderscore 1-hop} of FB15k-237 takes the form:
\begin{equation}
\label{eq:fw5}
\begin{split}
&h \| D_{h} \| r_{1} \| t_{1} \| D_{t_{1}} \| [SEP]\| ... \| [SEP] \| h \| D_{h} \| r_{n} \| \\
&t_{n} \| D_{t_{n}} \; \text{ or } h_{1} \| D_{h_{1}} \| r_{1} \| t \| D_{t} \| [SEP]\| ... \| [SEP] \| \\
&h_{n} \| D_{h_{n}} \| r_{n} \| t \| D_{t}
\end{split}
\end{equation}
The former is for the tail-masked version, and the latter is for the head-masked version.

\textbf{Stage 4 of retrieval process.} The \textbf{triple\textunderscore with\textunderscore $\boldsymbol{d}$} and \textbf{filtered\textunderscore 1-hop} are concatenated, and the $[CLS]$ token is added at the beginning of the sequence. The \textbf{triple\textunderscore with\textunderscore 1-hop} of WN18RR in Figure~\ref{fig:framework} takes the form:
\begin{equation}
\label{eq:fw6}
\begin{split}
&[CLS] \| [MASK] \| r \| t \| d_{t} \| [SEP] \| h_{1} \| r_{1} \| t \| \\
&[SEP]\| ... \| [SEP] \| h_{n} \| r_{n} \| t \; \text{ or } [CLS] \| h \| d_{h} \| \\
&r \| [MASK] \| [SEP] \| h \| r_{1} \| t_{1} \| [SEP]\| ... \\
&\| [SEP] \| h \| r_{n} \| t_{n}
\end{split}
\end{equation}

The \textbf{triple\textunderscore with\textunderscore 1-hop} of FB15k-237 takes the form:
\begin{equation}
\label{eq:fw7}
\begin{split}
&[CLS] \| [MASK] \| r \| t \| d_{t} \| [SEP] \| h \| D_{h} \| \\
&r_{1} \| t_{1} \| D_{t_{1}} \| [SEP]\| ... \| [SEP] \| h \| D_{h} \| r_{n} \| \\
&t_{n} \| D_{t_{n}} \; \text{ or } \; [CLS] \| h \| d_{h} \| r \| [MASK] \| \\
&[SEP] \| h_{1} \| D_{h_{1}} \| r_{1} \| t \| D_{t} \| [SEP]\| ... \\
&\| [SEP] \| h_{n} \| D_{h_{n}} \| r_{n} \| t \| D_{t}
\end{split}
\end{equation}
Finally, the input raw text, $x_{i} = \textbf{triple\textunderscore with\textunderscore 1-hop}$, is sent to the BERT model.

\subsection{Fine-tuning}

Since the vocabulary size of the BERT-base-cased tokenizer is $28,009$, which may not include all entities in the training data, we extend the vocabulary size of the tokenizer and embedding layer so that we can use a single $[MASK]$ token to mask out each entity during the fine-tuning stage. For WN18RR and FB15k-237, we increase the vocabulary size of the tokenizer and the dimension of the classification layer to $57,005$ and $42,516$, respectively. Then we restore the pre-training weight of BERT and train the extended version of the embedding layer. All training samples $x_{i}$, including the head-masked and tail-masked versions, are sent to the BERT model during the fine-tuning stage. Then the output embedding of the encoder's $[MASK]$ token is used as the input of the classification layer.

\subsection{Loss and Training Strategy}
The model with a large-dimension classification layer may yield low precision at $k$ if optimized by the binary cross entropy (BCE) loss. We propose a loss function and a training strategy to handle the issue. Our training strategy can be divided into three stages. The hyperparamters used in each stage can be found in table \ref{training_hyperparameter}, while the hyperparameters of the warmup scheduler are fixed for each stage as shown in equation \ref{eq:lr_rate}.

\textbf{Stage 1 of training.} Inspired by Focal loss~\cite{lin2017focal} and BAT~\cite{liu2022born}, a coefficient $\alpha$ is added to the BCE loss to encourage the model to predict $1$ instead of $0$ as follows:
\begin{equation}
\label{eq:loss1}
L_{stage1} = -\frac{1}{N} \sum\limits_{i=1}^N \alpha y_{i} (log\hat{y_{i}}) + (1-y_{i}) log(1-\hat{y_{i}})
\end{equation}
where $N$ denotes the sample size, $\alpha$ denotes the positive coefficient, $y_{i}$ denotes the true label, and $\hat{y_{i}}$ denotes the predicted probability. In this stage, we set $\alpha=30,000$ for both datasets so that the recall of the model on the testing set is large enough. Any $\alpha$ that enables the model to achieve a recall of $0.85$ or higher will work. Once the recall of the model on the testing set is high enough, we will proceed to the next stage to improve the precision.

\textbf{Stage 2 of training.} We try to raise the precision at $k$ by incorporateing the precision coefficint $p$ into equation~\ref{eq:loss1} as follows: 
\begin{equation}
\label{eq:loss2}
\begin{split}
L_{stage2} = -\frac{1}{N} \sum\limits_{i=1}^N \frac{1}{max(p, 0.01)}(\alpha y_{i} (log\hat{y_{i}}) \\
+ (1-y_{i}) log(1-\hat{y_{i}}))
\end{split}
\end{equation}
where $p$ is the precision of $\hat{y_{i}}$. At this stage, we lower the value of $\alpha$ for WN18RR to $100$ and that for FB15k-237 to $20$. Once the P@$1$ (precision at $1$) of the model on the testing set is high enough, we will proceed to the next stage. 

\textbf{Stage 3 of training.} We further improve performance by lowering the value of $alpha$. The loss is the same as equation~\ref{eq:loss2} while the value of $\alpha$ for both WN18RR and FB15k-237 is set to $2$.

\section{Experimental Study}
\subsection{Competing Models and Evaluation Measures}

The performance of the proposed method for the proposed task is compared with those of the models for the XMTC  (extreme multi-label text classification) task since the proposed task is similar to the XMTC task. We choose AttentionXML~\cite{you2019attentionxml} and LightXML~\cite{jiang2021lightxml} as our competing models. The P@k (precision at k) is selected as the evaluation metric for the reason that it is widely used for the XMTC task in the literature~\cite{yu2019x, jiang2021lightxml, you2019attentionxml}.

\subsection{Experimental Settings}

At each stage of training, we adopt a custom learning rate scheduler used in the original Transformer~\cite{vaswani2017attention} with a slight transformation:
\begin{equation}
\label{eq:lr_rate}
\begin{split}
\text{learning rate} = \frac{1}{2\sqrt{d_{model}}}min( \frac{1}{\sqrt{\text{step number}}}, \\
(\text{step number})(\text{warmup step}^{-1.5}))
\end{split}
\end{equation}
where $d_{model}$ denotes the dimension of the model which is set to $768$ and the warmup step is $12,000$. The values of the dropout ratio for the embedding layer and the self-attention are both $0.1$. The Automatic Mixed Precision (AMP) package is used to reduce GPU memory usage. The gradient accumulation technique is used to imitate a larger batch size. batch size. All experiments are conducted on the Tesla V100 SXM2 GPU. The rest of the hyperparameters in our experiments at each stage are shown in Table~\ref{training_hyperparameter}.

\begin{table}[]

\centering
\scalebox{0.8}{
\begin{tabular}{lrrrrrrrr}
\hline
\multirow{2}{*}{Datasets} & \multicolumn{2}{r}{stage1} & \multicolumn{2}{r}{stage2} & \multicolumn{2}{r}{stage3} & \multirow{2}{*}{$b$} & \multirow{2}{*}{$L_{t}$} \\ \cline{2-7}
                          & $B$          & $E$          & $B$          & $E$          & $B$          & $E$          &                      &                          \\ \hline
WN18RR                    & 160          & 31           & 160          & 9            & 160          & 5            & 768                  & 512                      \\
FB15k-237                 & 320          & 42           & 160          & 7            & 160          & 12           & 768                  & 512                      \\ \hline
\multicolumn{9}{c}{\makecell{$B$: batch size, $E$: \# epoch, $b$: dimension of embeddings,}} \\
\multicolumn{9}{c}{\makecell{$L_{t}$: maximum length of input tokens}} \\

\end{tabular}}
\caption{Hyperparameters used in our experiments at each training stage.}
\label{training_hyperparameter}
\end{table}
\subsection{Performance Comparison with State-of-the-art Models}

\begin{table}[]

\centering
\scalebox{0.7}{
\begin{tabular}{lrrrrrr}
\hline
\multicolumn{1}{l}{\multirow{2}{*}{Model}}   & \multicolumn{3}{c}{WN18RR}                                           & \multicolumn{3}{c}{FB15k-237}                  \\ \cmidrule{2-7} 
\multicolumn{1}{r}{}                          & P@1            & P@3            & \multicolumn{1}{r}{P@5}            & P@1            & P@3           & P@5            \\ \hline
\multicolumn{7}{l}{XMTC models with  $h \| r$  and  $r\|t$  as input}                                                                                         \\ \hline
\multicolumn{1}{l}{LightXML}                  & \multicolumn{1}{r}{21.52}          & \multicolumn{1}{r}{12.74}          & \multicolumn{1}{r}{9.26}           & 19.81          & 10.48         & 7.83           \\
\multicolumn{1}{l}{AttentionXML}              & 20.43          & 12.82          & 9.51           & 19.69          & 10.44         & 7.80            \\ \hline
\multicolumn{7}{l}{XMTC models with $h \| d_{h} \| r$ and $r \| t \| d_{t}$ as input}                                                                  \\ \hline
\multicolumn{1}{l}{LightXML}                  & 31.84          & 17.41          & \multicolumn{1}{c}{12.03}          & 23.57          & 12.42         & 9.16           \\
\multicolumn{1}{l}{AttentionXML}              & 25.55          & 16.31          & 11.53          & 21.70           & 11.58         & 8.57           \\ \hline
\multicolumn{1}{l}{Our Model} & \textbf{35.43}           & \textbf{26.24}           & \textbf{23.27}         & \textbf{32.25} & \textbf{22.70} & \textbf{19.79} \\ \hline
\multicolumn{7}{c}{\makecell{$h$: head entity, $r$: relation, $t$: tail entity, $d_{h}$: description of }} \\
\multicolumn{7}{c}{\makecell{the given head entity in an incomplete triple, $d_{t}$: description }} \\
\multicolumn{7}{c}{\makecell{of the given tail entity in an incomplete triple, $\|$: concatenate}} \\
\multicolumn{7}{c}{\makecell{ }} \\
\end{tabular}}
\caption{Performance comparison with competing models.}
\label{main_result}

\end{table} 

For  the AttentionXML~\cite{you2019attentionxml} and LightXML~\cite{jiang2021lightxml} models, we first train the models with raw text input $x_{i} = h\|r \; \text{and} \; r \| t$, where $\|$ denotes the concatenate operation. We then concatenate the description data with the raw text input $x_{i} = h \| d_{h} \| r$ \text{and} $r \| t \| d_{t}$ and train the models. Table~\ref{main_result} shows that our model yields significantly higher precision at $k$ ($k = 1, 3, 5$) than the two competing models on both datasets. The improvement of the performance of the competing models also proves the effectiveness of our proposed FB15k-237's description dataset and the WN18RR's short description dataset.

\begin{table}[]

\centering
\scalebox{0.6}{
    \begin{tabular}{cll}
    \hline
        & WN18RR & FB15k-237  \\ \hline
    augmenting &\multirow{2}{*}{$Z$ + $d$ + 512 + SimCSE} & \multirow{2}{*}{$Z$ + $d$ + top 3 + SimCSE + $D$} \\
    strategy & & \\
    (P@1, P@3, P@5) & \textbf{(35.43, 26.24, 23.27)} & \textbf{(32.25, 22.7, 19.79)} \\ \hline
    \multirow{3}{*}{exp. 1} &  ${-}$ SimCSE $+$ random	& $-$ SimCSE $+$ random \\ 
       & (0.02, 0.01, 0.01) & (30.91$\pm$0.41, 20.26$\pm$0.31, \\
       & &  16.97$\pm$0.27)\\ \hline
      \multirow{2}{*}{exp. 2}  &   ${-}$ 512 ${+}$ top 3 $+$ $D$ & ${-}$ $D$ \\ 
       & (0.02, 0.01, 0.01) & (28.96, 19.23, 16.23) \\ \hline
        \multirow{2}{*}{exp. 3} &  ${-}$ 512 ${+}$ top 3 & ${-}$ top3 + 512  \\ 
       & (0.02, 0.01, 0.01) & (26.56, 16.80, 13.96) \\ \hline
       \multirow{2}{*}{exp. 4}  &  ${-}$ 1-hop	& ${-}$ 1-hop \\ 
       & (0.04, 0.02, 0.02) & (30.82, 20.39, 17.14) \\ \hline
        \multirow{2}{*}{exp. 5} &  ${-}$ $d$ & ${-}$ $d$	\\ 
       & (0.00, 0.01, 0.01) & (29.44, 19.66, 16.71) \\ \hline
       \multicolumn{3}{c}{\makecell{$Z$: incomplete triple, $d$: description of a given node in an incomplete triple,}}\\
       \multicolumn{3}{c}{\makecell{ $D$: description of entities in one-hop neighborhood, 512: sample one-hop }}\\
       \multicolumn{3}{c}{\makecell{triples randomly from one-hop neighborhood such that the number of input}}\\
       \multicolumn{3}{c}{\makecell{tokens doesn't exceed 512, SimCSE: employ the model for similarities }} \\
       \multicolumn{3}{c}{\makecell{calculation, top3: select top 3 one-hop triples by SimCSE with highest }} \\
        \multicolumn{3}{c}{\makecell{similarity scores.}}\\
    
    \end{tabular}}
\caption{Ablation study of different input formats for the BERT model.}
\label{ablation_input} 
\end{table}

\begin{table*}

\centering
\scalebox{0.6}{
\begin{tabular}{lcccccc}
\hline
 & \multicolumn{3}{c}{\textbf{WN18RR}} & \multicolumn{3}{c}{\textbf{FB15k-237}}\\ \hline\hline
    & stage1 ( w/o $p$ in loss) & stage2 ( w/ $p$ in loss) & stage3 ( w/ $p$ in loss) 
 &  stage1 ( w/o $p$ in loss) & stage2 ( w/ $p$ in loss) & stage3 ( w/ $p$ in loss) \\ \hline
{\multirow{2}{*}{strategy1}}&                     &                       & $\alpha=2$ &                      &                       & $\alpha=\textbf{2}$     \\ 
                            &                     & $\alpha=20$           & (34.93, 26.06, 23.18) &                      & $\alpha=\textbf{20}$           &  \textbf{(32.25, 22.70, 19.79)}\\ \cline{1-1} \cline{4-4} \cline{7-7}
{\multirow{2}{*}{strategy2}}&                     & (32.59, 21.56, 17.89) & $\alpha=5$ &                      & \textbf{(24.58, 14.57, 11.47)} & $\alpha=5$ \\
                            & $\alpha=30000$  &                       & (35.03, 25.41, 22.53) &   $\alpha=30000$       &                       & ( 26.00, 16.21, 13.26) \\ \cline{1-1} \cline{3-4} \cline{6-7}
{\multirow{2}{*}{strategy3}}&       &                       & $\alpha=2$ &        &                       & $\alpha=2$\\
                            & (25.68, 16.07, 11.73) & $\alpha=50$           & (35.08, 26.33, 23.18) & (22.30, 12.61, 9.65) & $\alpha=50$           & (29.05, 19.67, 16.77)\\ \cline{1-1} \cline{4-4} \cline{7-7}
{\multirow{2}{*}{strategy4}}& & (32.63, 20.57, 16.36) & $\alpha=5$ &   & (25.03, 14.46, 11.15) & $\alpha=5$\\
                            & recall = 0.855 &                       & (34.73, 24.98, 21.85) &  recall = 0.957                   &                       & (26.75, 16.55, 13.46) \\ \cline{1-1} \cline{3-4} \cline{6-7}
{\multirow{2}{*}{strategy5}}&                     &                       & $\alpha=\textbf{2}$ &                     &                       & $\alpha=2$\\
                            &                     & $\alpha=\textbf{100}$          & \textbf{(35.43, 26.24, 23.27)} &                      & $\alpha=100$          & (29.75, 19.54, 16.42)\\ \cline{1-1} \cline{4-4} \cline{7-7}
{\multirow{2}{*}{strategy6}}&                     & \textbf{(32.68, 20.26, 15.90)} &  $\alpha=5$ &                      & (24.93, 14.24, 11.01) &  $\alpha=5$\\
                            &                     &                       & (35.04, 25.47, 22.26) &                     &                       & (25.94, 16.24, 13.23) \\ \hline
\multicolumn{7}{c}{} \\
\end{tabular}}
\caption{Ablation study for different values of $\alpha$ in the loss function at each stage.}
\label{alpha_ablation}

\end{table*}

\begin{table*}

  \label{sample-table}
  \centering
  \scalebox{0.65}{
 \begin{tabular}{lcccccc}
\hline
 & \multicolumn{3}{c}{\textbf{WN18RR}} & \multicolumn{3}{c}{\textbf{FB15k-237}}\\ \hline\hline
  & stage1 ($\alpha=30000$) & stage2 ($\alpha=100$) & stage3 ($\alpha=2$) & stage1 ($\alpha=30000$) & stage2 ($\alpha=20$) & stage3 ($\alpha=2$)\\ \hline
 {\multirow{2}{*}{strategy1}} &                &                 & w/o $p$ in loss &                       &                       & w/o $p$ in loss \\
 &                 & w/o $p$ in loss & (34.75, 26.04, 22.91) &                       & w/o $p$ in loss        & (28.22, 18.18, 15.20) \\ \cline{1-1} \cline{4-4} \cline{7-7}
{\multirow{2}{*}{strategy2}} & w/o $p$ in loss & (33.04, 20.53, 16.31) & w/ $p$ in loss & w/o $p$ in loss       & (25.03, 14.72, 11.52) & w/ $p$ in loss\\
 & (25.68, 16.07, 11.73) &                       & (34.76, 26.52, 23.38) & (22.30, 12.61, 9.65) &                         &  (28.72, 18.73, 15.79)\\ \cline{1-1} \cline{3-4} \cline{6-7}
{\multirow{2}{*}{strategy3}} &                 & w/ $p$ in loss & w/ $p$ in loss &                       & w/ $p$ in loss         & w/ $p$ in loss \\
 &                  & (32.68, 20.26, 15.90) &  \textbf{(35.43, 26.24, 23.27)} &                       & (24.58, 14.57, 11.47)  &   \textbf{(32.25, 22.70, 19.79)}\\ \hline
\end{tabular}}
\caption{Ablation study for loss function on $p$ at each stage.}
\label{precision_ablation}

\end{table*}

\section{Ablation Study}
\subsection{Ablation Study for Our Framework}

As shown in Figure~\ref{fig:framework}, we view our input raw text (\textbf{triple\textunderscore with\textunderscore 1-hop}) as two different parts; one of them is the main triple (\textbf{triple\textunderscore with\textunderscore $\boldsymbol{d}$}) while the other is the one-hop neighbors (\textbf{filtered\textunderscore 1-hop}). Several different components of our framework are removed or replaced in the ablation study: (1) remove $d$ (description in \textbf{triple\textunderscore with\textunderscore $\boldsymbol{d}$}), (2) replace the SimCSE model with random sampling, (3) remove 1-hop (\textbf{filtered\textunderscore 1-hop}), (4) change the filtering strategy of \textbf{filtered\textunderscore 1-hop} from $top \; 3$ (top three triples) to $512$ (input token count is limited to 512 as mentioned in section~\ref{sec_framework}), (5) remove or add $D$ (the descriptions in \textbf{filtered 1-hop}). Table~\ref{ablation_input} shows that $d$ and the SimCSE model are essential to our framework, while $D$ is essential to FB15k-237. Furthermore, from the performances of all experiments of WN18RR and experiments $2, 3$ of FB15k-237, we can observe that the choice of \textbf{filtered\textunderscore 1-hop} is also critical. Different choices of \textbf{filtered\textunderscore 1-hop} can lead to significant differences in performance. Note that the error bar in experiment $1$ of FB15k-237 is the average error bar for performing the experiments six times, each with a random seed.

\subsection{Ablation Study for Our Loss and Training Strategy}

To demonstrate the effectiveness of $\alpha$ (positive coefficient) in the proposed loss, we test different values of $\alpha$ at each stage as shown in Table~\ref{alpha_ablation}. At the first stage, our intuition to set $\alpha=30,000$ is to increase the tendency for the model to predict $1$ instead of $0$ and raise the recall value at stage one. Once the recall is high enough, we can gradually raise the precision by lowering the value of $\alpha$ and adding precision to the loss function. On the contrary, if we set $\alpha=1$ without multiple training stages, the precision at $k$, $k=1,2,3$, are zero. As the choice of $\alpha$ at stage two, we prefer a larger positive value ($\alpha=100$) for the model trained on WN18RR, whereas a smaller value ($\alpha=20$) is preferred for the model trained on FB15k-237. We believe that the causes of this difference are: (1) As mentioned in section~\ref{datasets}, compared with FB15k-237, it's relatively difficult for a model to infer from a given node to another for WN18RR. (2) the recall of the model in stage one of WN18RR is smaller than that of FB15k-237. A larger $\alpha$ enables the tendency for the model to make a positive prediction which will have a greater chance of making a good guess. At the last stage, a small value of $\alpha$ both is preferred for the two datasets such that the model can achieve a higher precision.

To demonstrate the effectiveness of $p$ (precision) in loss, we remove the $p$ at training stages two and three as shown in Table~\ref{precision_ablation}. The table shows that the performance of the model trained with precision in loss is improved by more than three percent on FB15k-237 compared with that of the model trained on loss without precision factor.

\subsection{Limitation}

Our proposed loss function and the corresponding training strategy are designed for our proposed task. Although it has decent performance on the extreme multi-label knowledge graph link prediction task, it may not be able to generalize to other tasks without further experiments.

\section{Conclusion}

Our experiments demonstrate that knowledge graphs with different characteristics require different augmenting strategies. WN18RR has two characteristics: (1) a relatively large proportion of disconnected triples and (2) a low number of one-hop neighbors per node. These issues raise the difficulty for the model to infer from a given node to another on WN18RR. To alleviate these issues, we augment the input of BERT by corresponding description data and the one-hop neighbors chosen from our retriever as additional information.

On the other hand, FB15k-237 possesses two characteristics: (1) neighborhoods in FB15k-237 are typically large, and (2) most entities in FB15k-237 are proper nouns. Descriptions corresponding to nodes play a critical role when the given node and the target nodes are all proper nouns. This additional information helps the model to understand these proper nouns. Therefore, we add the descriptions corresponding to every entity when making inferences in FB15k-237, while WN18RR only needs to add the description in the main triple. Moreover, when inferring multiple nodes that are proper nouns, information with relatively low similarity can be viewed as noises. These noises can seriously lower the task performance. To handle this problem, the retriever chooses the top three most relevant one-hop neighbors and their descriptions to augment the input of BERT.

For both knowledge graphs, we find it difficult for the model to converge to a desired local minimum when optimized by binary cross-entropy loss. This is caused by the large label space, which leads to an extremely high-dimensional classification layer. To address the issue, our proposed loss and training strategy encourages the model to predict 1 for every class in the first stage by introducing a large positive coefficient. In the second stage, we incorporate the precision coefficient in the loss used in the first stage to restrict the model for guessing 1 and raise the precision. In our last stage, we achieve the desired precision at $k$ by reducing the influence of the positive coefficient on the model.

\medskip
{
\small
\bibliographystyle{acl_natbib.bst}
\bibliography{ref.bib}
}

\medskip

%%%%%%%%%%%%%%%%%%%%%%%%%%%%%%%%%%%%%%%%%%%%%%%%%%%%%%%%%%%%

\end{document}